\documentclass[11pt,a4paper]{article}
\pdfoutput=1
\usepackage[hyperref]{acl2021}
\usepackage{times}
\usepackage{latexsym}
\usepackage{comment}
\usepackage{amsmath}
\usepackage{amssymb}
\usepackage{lipsum}  
\newcommand\HH{\mathbb H}
\usepackage{footnote}

\makeatletter
\newcommand\footnoteref[1]{\protected@xdef\@thefnmark{\ref{#1}}\@footnotemark}
\makeatother

\usepackage{tikz}
\usetikzlibrary{intersections, calc, angles}

\usepackage{pifont}
\usepackage{microtype}

\aclfinalcopy 
\def\aclpaperid{2720} 

\newcommand\BibTeX{B\textsc{ib}\TeX}

\title{Hyperbolic Temporal Knowledge Graph Embeddings with Relational and Time Curvatures}

\author{Sebastien Montella \\
  Orange Labs, Lannion \\
  Aix Marseille Univ., CNRS, LIS, Marseille, France \\
  \texttt{sebastien.montella@orange.com} \\\AND
  
    Lina Rojas-Barahona \\
    Orange Labs, Lannion \\
  \\\And
    Johannes Heinecke \\
    Orange Labs, Lannion \\
    
    \texttt{\{linamaria.rojasbarahona, johannes.heinecke\}@orange.com~~~~~~~~~~~~~~~~~~~~~~~~~~~~~}
}
\date{}

\begin{document}
\maketitle
\begin{abstract}
Knowledge Graph (KG) completion has been excessively studied with a massive number of models proposed for the Link Prediction (LP) task. The main limitation of such models is their insensitivity to time. 
Indeed, the temporal aspect of stored facts is often ignored. 
To this end, more and more works consider time as a parameter to complete KGs. 
In this paper, we first demonstrate that, by simply increasing the number of negative samples, the recent \textsc{AttH} model can achieve competitive or even better performance than the state-of-the-art on Temporal KGs (TKGs), albeit its nontemporality. 
We further propose \textsc{Hercules}, a \textit{time-aware} extension of \textsc{AttH} model, which defines the curvature of a Riemannian manifold as the product of both relation and time. 
Our experiments show that both \textsc{Hercules} and \textsc{AttH} achieve competitive or new state-of-the-art performances on ICEWS04 and ICEWS05-15 datasets. Therefore, one should raise awareness when learning TKGs representations to identify whether time truly boosts performances.
\end{abstract}

\section{Introduction}
The prevalent manner to store factual information is the $\langle$\textit{s, p, o}$\rangle$ triple data structure where \textit{s}, \textit{p} and \textit{o} stand for the \textit{subject}, \textit{predicate} and \textit{object} respectively. An \textit{entity} denotes whether a subject or an object while a \textit{relation} denote a predicate that links two entities. A collection of triples defines a Knowledge Graph (KG) noted $\mathcal{G}(\mathcal{E}, \mathcal{R})$ with $\mathcal{E}$ the set of entities, \textit{i.e.} subjects and objects, corresponding to the nodes in the graph and $\mathcal{R}$ the set of predicates corresponding to directed edges. Shedding light on the type of connections between entities, KGs are powerful to work with for numerous downstream tasks such as question-answering \cite{bordes-etal-2014-question, hao-etal-2017-end, saxena-etal-2020-improving}, recommendation system \cite{yu-et-al-2014-personalized-recommender, zhang-etal-2016-recommender, Zhou_Liu_Liu_Liu_Gao_2017}, information retrieval \cite{lao-et-al-2010-rel-retrieval-kg, rocktaschel-etal-2015-injecting, xiong-et-al-2017-ir-kg}, or reasoning \cite{xian-et-al-2019-rl-reasoning, chen-et-al-2020-review-reasoning}.
However, KGs are sometimes incomplete and part of the knowledge is missing.
A major concern was therefore raised to predict missing connections between entities, stimulating research on the Link Prediction (LP) task. Intuition is to map each entity and relation into a vector space to learn low-dimensional embeddings such that valid triples maximize a defined scoring function and that fallacious triples minimize it. An approach is efficient if it can model multiple relational patterns. Some predicates are symmetric (\textit{e.g. marriedTo}), asymmetric (\textit{e.g. fatherOf}), an inversion of another relation (\textit{e.g. fatherOf} and \textit{childOf}) or a composition (\textit{e.g. grandfatherOf}). Distinct strategies were introduced by explicitly model those patterns \cite{bordes-et-al-2013-transe, yang-et-al-2015-distmult, trouillon-et-al-2016-complex, sun-et-al-2019-rotate}. However, hierarchical relations have remained challenging to model in Euclidean space. As demonstrated in  \citet{sarkar-et-al-2011-distortion}, tree structures are better embedded in hyperbolic spaces. 
Thus, hyperbolic geometry reveals to be a strong asset to capture hierarchical patterns. 
Nevertheless, the aforementioned approaches represent embeddings as invariant to time. For example, while writing this article, the triple $\langle$\textit{Donald Trump, presidentOf, U.S.}$\rangle$ is correct but will be erroneous at reading time due to the meantime United States presidential inauguration of Joe Biden. 
To address this issue, recent works considered using quadruplet written as $\langle$\textit{s, p, o, t}$\rangle$ by adding a time parameter $t$. Then, we note a Temporal Knowledge Graph (TKG) as $\mathcal{G}(\mathcal{E}, \mathcal{R}, \mathcal{T})$ with $\mathcal{T}$ the set of timestamps. Stating a precise time can be advantageous for diverse applications (disambiguation, reasoning, natural language generation, etc). Recent works toward TKG representations 
are essentially extensions of existing timeless KG embeddings 
that incorporate the time parameter in the computation of their scoring function. 

Similar to our work, \citet{han-etal-2020-dyernie} developed \textsc{DyERNIE}, an hyperbolic-based model inspired from \textsc{MuRP} \cite{balazevic-et-al-2019-murp}. \textsc{DyERNIE} uses a product of manifolds and adds a (learned) Euclidean time-varying representation for each entity such that each entity further possesses an \textit{entity-specific velocity vector} along with a static (\textit{i.e.} time-unaware) embedding.

In this paper, we first demonstrate that an optimized number of negative samples enables the \textsc{AttH} model \cite{chami-etal-2020-low} to reach competitive or new state-of-the-art performance on temporal link prediction while being unaware of the temporal aspect. We further introduce \textsc{Hercules}\footnote{\underline{\textbf{H}}yperbolic Representation with Tim\underline{\textbf{E}} and \underline{\textbf{R}}elational \underline{\textbf{CU}}rvatures for Tempora\underline{\textbf{L}} Knowledg\underline{\textbf{E}} Graph\underline{\textbf{S}}}, an extension of \textsc{AttH}. \textsc{Hercules} differs from \textsc{DyERNIE} in that: 
\begin{itemize}
\item Following \citet{chami-etal-2020-low}, we utilize Givens transformations and hyperbolic attention to model different relation patterns.
\item A single manifold is used. 
\item Curvature of the manifold is defined as the product of both relation and time parameters.
\end{itemize}
To the best of our knowledge, this is the first attempt to leverage the curvature of a manifold to coerce time-aware representation. We also provide an ablation study of distinct curvature definitions to investigate the surprising yet compelling results of \textsc{AttH} over time-aware models.


\section{Related Work}
In this section, we present existing methods on both KG and TKG vector representation.

\subsection{Timeless Graph Embeddings}
Previous works on KG completion essentially focused on undated facts with the $\langle$\textit{s, p, o}$\rangle$ formalism. \citet{bordes-et-al-2013-transe} initially proposed the \textsc{TransE} model considering the relation $p$ as a translation between entities in the embedding space. 
Several variants were then designed. \textsc{TransH} \cite{wang-et-al-2014-transh} adds an intermediate projection onto a relation-specific hyper-plane while \textsc{TransR} \cite{lin-et-al-2015-transr} maps entities to a relation-specific space of lower rank. However, translation-based approaches cannot model symmetric relations. \textsc{DistMult} \cite{yang-et-al-2015-distmult} solves this issue by learning a bilinear objective that attributes same scores to $\langle$\textit{s, p, o}$\rangle$ and $\langle$\textit{o, p, s}$\rangle$ triples. 
\textsc{ComplEx} \citet{trouillon-et-al-2016-complex} subsequently came up with complex embeddings. 
Tensor factorization techniques were also proposed. \textsc{RESCAL} \cite{nickel-et-al-2013-rescal} applies a three-way tensor factorization. \textsc{TuckER} \cite{balazevic-et-al-2019-tucker} uses \textit{Tucker decomposition} and demonstrates that \textsc{TuckER} is a generalization of previous linear models.
More recently, \textsc{RotatE} \cite{sun-et-al-2019-rotate} considered relations as rotations in a complex vector space which can represent symmetric relations as a rotation of $\pi$. \textsc{QuatE} \cite{zhang-et-al-2019-quaternion} further generalizes rotations using quaternions, known as \textit{hypercomplex} numbers ($\mathbb{C} \subset \HH$). A key advantage of quaternions is its non-commutative property allowing more flexibility to model patterns. Nonetheless, memory-wise, \textsc{QuatE} requires 4 embeddings for each entity and relation. 
To this extent, hyperbolic geometry provides an outstanding framework to produce shallow embeddings with striking expressiveness \cite{sarkar-et-al-2011-distortion, nickel-et-al-2017-poincare}. Both \textsc{MuRP} \cite{balazevic-et-al-2019-murp} and \textsc{AttH} \cite{chami-etal-2020-low} learn hyperbolic embeddings on a \emph{n}-dimensional Poincar\'e ball. Different to \textsc{MuRP},  \textsc{AttH} uses a trainable curvature for each relation. Indeed, \citet{chami-etal-2020-low} have shown that fixing the curvature of the manifold can jeopardize the quality of the returned embeddings. Therefore, defining a parametric curvature for a given relation helps to learn the best underlying geometry. 


\subsection{Time-Aware Graph Embeddings}
The above-mentioned techniques nevertheless disregard the temporal aspect. Indeed, lets consider the two following quadruplets $\langle$\textit{Barack Obama, visits, France, 2009-03-11}$\rangle$ and  $\langle$\textit{Barack Obama, visits, France, 2014-04-21}$\rangle$. Non-temporal models would exhibit the same score for both facts. However, the second quadruplet is invalid\footnote{In fact, Barack Obama visited Japan on April 21, 2014, not France} and should therefore get a lower score. For this reason, several works contributed to obtain time-aware embeddings. Thanks to the existing advancements on graph representations, many strategies are straightforward extensions of static approaches. \textsc{TTransE} \cite{leblay-et-al-2018-ttranse} alters the scoring function of \textsc{TransE} to encompass time-related operations such as time translations. Likewise, \textsc{TA-TransE} \cite{garcia-duran-etal-2018-learning} uses LSTMs \cite{hochreiter-et-al-1997-lstm} to encode a temporal predicate which carries the time feature. By analogy with \textsc{TransH}, \textsc{HyTe} \cite{dasgupta-etal-2018-hyte} learns time-specific hyper-planes on which both entities and relations are projected. Then, another family of temporal extensions are derived from \textsc{DistMult} such as 
\textsc{Know-Evolve} \cite{trivedi-et-al-2017-knowevolve}, \textsc{TA-DistMult} \cite{garcia-duran-etal-2018-learning}, or \textsc{TDistMult} \cite{ma-et-al-2019-tdistmult} that also utilize a bilinear scoring function. \textsc{DE-SimplE} \cite{goel-et-al-2020-diachronic} provides diachronic entity embeddings inspired from diachronic word embeddings \cite{hamilton-etal-2016-diachronic}. Recently, \textsc{ATISE} \cite{xu-et-al-2019-atise} embeds entities and relations as a multi-dimensional Gaussian distributions which are time-sensitive. An advantage of \textsc{ATISE} is its ability to represent time uncertainty as the covariance of the Gaussian distributions. \textsc{TeRo} \cite{xu-et-al-2020-tero} combines ideas from \textsc{TransE} and \textsc{RotatE}. It defines relations as translations and timestamps as rotations. As far as we are aware, \textsc{DyERNIE} \cite{han-etal-2020-dyernie} is the first work to contribute to hyperbolic embeddings for TKG. It achieves state-of-the-art performances on the benchmark datasets ICEWS14 and ICEWS05-15. Time is defined as a translation on a product of manifolds with trainable curvatures using a \textit{velocity vector} for each entity.

In our work, we demonstrate that using a single manifold with learnable relational and time curvatures is sufficient to reach competitive or new state-of-the-art performances.

\section{Problem Definition}
Lets consider a valid quadruplet $\langle$\textit{s, p, o, t}$\rangle$ $\in \mathcal{S} \subset \mathcal{E}\times\mathcal{R}\times\mathcal{E}\times\mathcal{T}$, with $\mathcal{E}$, $\mathcal{R}$ and $\mathcal{T}$ the sets of entities, relations and timestamps respectively and $\mathcal{S}$ the set of correct facts. A scoring function $f:\mathcal{E}\times\mathcal{R}\times\mathcal{E}\times\mathcal{T} \rightarrow \mathbb{R}$ is defined such that $f(s,p,o,t)$ is maximized for any quadruplet $\in$ $\mathcal{S}$, and minimized for corrupted quadruplet ($\notin$ $\mathcal{S}$). Throughout the optimization of the foregoing constraint, representations of entities, relations and times are learned accordingly. The resulting embeddings should then capture the multi-relational graph structure. Thus, $f$ is gauging the probability that an entity \textit{s} is connected to an entity \textit{o} by the relation \textit{p} at time \textit{t}. 

\section{Hyperbolic Geometry}
\begin{figure}[h]
    \centering
    \begin{tikzpicture}[
  point/.style = {draw, circle, fill=black, inner sep=0.7pt},
]
\def\rad{2cm}
\definecolor{manifold_color}{rgb}{0.80,0.98,0.98}
\coordinate (O) at (0,0); 
\coordinate (N) at (0,\rad); 

\filldraw[ball color=manifold_color] (O) circle [radius=\rad];
\draw[dashed] 
  (\rad,0) arc [start angle=0,end angle=180,x radius=\rad,y radius=5mm];
\draw
  (\rad,0) arc [start angle=0,end angle=-180,x radius=\rad,y radius=5mm];
\begin{scope}[xslant=0.5,yshift=\rad,xshift=-2]
\filldraw[fill=gray!50,opacity=0.3]
  (-3,1) -- (3,1) -- (3,-1) -- (-3,-1) -- cycle;

  (-4,1) -- (3,1) -- (3,-1) -- (-4,-1) -- cycle;
\node at (2,0.6) {$\mathcal{T}_{x}^{c}\mathbb{B}^{n, c}$};  
\node at (1.2 * \rad,-2 * \rad) {$\mathbb{B}^{n, c}$};
\end{scope}
\draw[color=red, ->, thick] (0, \rad) -- (2.3, 1.4);
\node at (1.25,1.9) {$u$};

\draw[dotted, color=blue, thick, ->] (2.3, 1.4) arc (5:-50:1.5) coordinate (c);
\node[text=blue] at ($(c) + (-0.75,-0.06)$) {$\exp_{x}^{c}(\textcolor{black}{u})$} ;

\draw[dotted, color=red, thick, ->] ($ (c) + (0.3,0) $) arc (-50:8:1.35) coordinate (d);
\node[text=red] at ($(d) + (0.6,-0.8)$) {$\log_{x}^{c}(\textcolor{black}{v})$} ;

\draw[dashed, color=blue, thick] (0, \rad) arc (250:215:-2.8) coordinate (e) ;
\draw[solid, color=blue, thick, ->] (e) arc (215:198:-2.8) ;
\node at ($(e) + (-0.25, -0.2)$) {$v$};

\draw[dashed]
  (N) node[above] {$x$} -- (O) node[below] {$O$};
\node[point] at (N) {};
\end{tikzpicture}
    \caption{Illustration of the exponential and logarithmic maps between the Poincar\'e ball $\mathbb{B}^{n, c}$ and the tangent space  $\mathcal{T}_{x}^{c}\mathbb{B}^{n, c}$.}
    \label{fig:poincare_ball}
\end{figure}
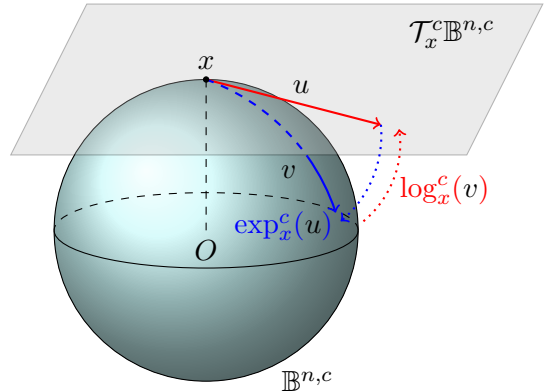

Hyperbolic geometry belongs to non-Euclidean geometry. In contrast to Euclidean geometry relying on Euclid's axioms \cite{euclid-et-al-1956-element}, non-Euclidean geometry rejects the fifth axiom known as the parallel postulate. It states that given a point $x$ and a line $l_{1}$, there exists a unique line $l_{2}$ parallel to $l_{1}$ passing through $x$. This is only possible due to a (constant) zero curvature of the space. The curvature defines how much the geometry differs from being flat. The higher the absolute curvature, the curvier. Euclidean space has a zero curvature hence called \textit{flat space}. When represented in an Euclidean space, straight lines become curved, termed as \textit{geodesics} (Fig. \ref{fig:poincare_ball}).

Hyperbolic geometry comes with a constant negative curvature. In our study, as \citet{nickel-et-al-2017-poincare, han-etal-2020-dyernie, chami-etal-2020-low}, we make use of the Poincar\'e ball ($\mathbb{B}^{n, c}$, $g^{\mathbb{B}}$) which is a \textit{n}-dimensional Riemannian manifold $\mathbb{B}^{n,\:c} = \{x \in \mathbb{R}^{n}:\lVert x \rVert^{2}<\frac{1}{c}\}$ of constant curvature $-c$ ($c>0$) equipped with Riemannian metric $g^{\mathbb{B}}$ and where $\lVert . \rVert$ denotes the $L_{2}$ norm. The metric tensor $g^{\mathbb{B}}$ gives information about how distances should be computed on the manifold and is defined as $g^{\mathbb{B}}=(\lambda_{x}^{c})^{2}I_{n}$ with the \textit{conformal factor} $\lambda_{x}^{c}=\frac{2}{1-c\lVert x \rVert^{2}}$ and $I_{n}$ the identity matrix of size $n \times n$. We write $\mathcal{T}_{x}^{c}\mathbb{B}^{n, c}$ the \textit{n}-dimensional tangent space at $x \in \mathbb{B}^{n, c}$. At the difference of $\mathbb{B}^{n, c}$, $\mathcal{T}_{x}^{c}\mathbb{B}^{n, c}$ locally follows an Euclidean geometry. As illustrated in Fig. \ref{fig:poincare_ball}, we can project $v \in \mathbb{B}^{n, c}$ on $\mathcal{T}_{x}^{c}\mathbb{B}^{n, c}$ at $x$ via the logarithmic map. Inversely, we can map a point $u \in \mathcal{T}_{x}^{c}\mathbb{B}^{n, c}$ on $\mathbb{B}^{n, c}$ at $x$ via the exponential map. A closed-form of those mappings exist when $x$ corresponds to the origin (Eqs. \ref{eq:expmap} and \ref{eq:logmap}).
\begin{equation}
    \exp_\textbf{0}^{c}(u) = \tanh(\sqrt{c}\lVert u \rVert)\frac{u}{\sqrt{c}\lVert u \rVert}
    \label{eq:expmap}
\end{equation}

\begin{equation}
    \log_\textbf{0}^{c}(v) = \tanh^{-1}(\sqrt{c}\lVert v \rVert)\frac{v}{\sqrt{c}\lVert v \rVert}
    \label{eq:logmap}
\end{equation}
Contrary to $\mathcal{T}_{x}^{c}\mathbb{B}^{n, c}$, the Euclidean addition on the hyperbolic manifold does not hold. Alternately, we use the M\"obius addition \cite{vermeer-et-al-2005-ungar, ganea-et-al-2018-hnn}  satisfying the boundaries constraints of the manifold. It is however non-commutative and non-associative. The closed-form is presented in Eq. \ref{eq:mobius_add}.
\begin{equation}
\begin{split}
x&\oplus_{c}y = \\ & \frac{(1-2cx^{T}y-c\lVert y \rVert^{2})x + (1+c \lVert x \rVert ^{2})y}{1-2cx^{T}y + c^{2}\lVert y \rVert^{2}\lVert x \rVert ^{2}}
\end{split}
\label{eq:mobius_add}
\end{equation}
The distance between two points $x$ and $y$ on $\mathbb{B}^{n, c}$ is the hyperbolic distance $d_{\mathbb{B}^{n, c}}(x, y)$ defined as:
\begin{equation}
    d_{\mathbb{B}^{n, c}}(x, y)=\frac{2}{\sqrt{c}}\tanh^{-1}(\sqrt{c}\lVert -x\oplus_{c}y\rVert)
\end{equation}


\section{From \textsc{AttH} to \textsc{Hercules}}

Given a quadruplet, $\langle$\textit{s, p, o, t}$\rangle$, we note $e_{s}^{H}$, $r_{p}^{H}$ and $e_{o}^{H}$ the hyperbolic embeddings of the subject, predicate and object respectively.\footnote{Since \textsc{AttH} is not considering time, the parameter $t$ is not used.} \textsc{AttH} uses relation-specific embeddings, rotations, reflections and curvatures. The curvature is defined as depending on the corresponding relation \textit{p} involved. Precisely, a relation \textit{p} is attributed with an individual parametric curvature $c_{p}$. The curvature $c_{p}$ is defined in Eq. \ref{eq:curvature_definition} as:
\begin{equation}
\label{eq:curvature_definition}
    c_p = \sigma(\mu_p)
\end{equation}
where $\mu_p$ is a trainable parameter $ \in \mathbb{R}$ and $\sigma$ is a smooth approximation of the ReLU activation function defined in $[0, +\infty]$.
With such approach, the geometry of the manifold is learned, thus modified for a particular predicate. The curvature dictates how the manifold is shaped. Changing the curvature of the manifold implies changing the positions of projected points. 
This means that for distinct relations, the same entity will have different positions because of the different resulting geometries for each relation. For example, lets consider the triples $t_1:=\langle$\textit{Barack Obama, visit, France}$\rangle$ and $t_2 := \langle$\textit{Barack Obama, cooperate, France}$\rangle$. The Euclidean representations of entities \textit{Barack Obama} and \textit{France} from both facts will be projected onto the riemannian manifold. However, the structure (i.e. curvature) of the manifold changes as a function of the relation of each fact (\textit{i.e.} '\textit{visit}' and '\textit{cooperate}'). Therefore, the resulting hyperbolic embbeding of \textit{Barack Obama} of $t_1$ will not be the same resulting hyperbolic embedding of \textit{Barack Obama} in $t_2$. By analogy, the same holds for entity \textit{France}. 

In order to learn rotations and reflections, \textsc{AttH} uses 2 $\times$ 2 Givens transformations matrices. Those transformations conserve relative distances in hyperbolic space and can therefore directly be applied to hyperbolic embeddings (isometries). We note $W^{rot}_{\Theta_{p}}$ and $W^{ref}_{\Phi_{p}}$ the block-diagonal matrices where each element on their diagonals is given by $G^{+}(\theta_{p, i})$ and $G^{-}(\phi_{p, i})$ respectively, with $i$ the $i^{th}$ element of the diagonal (Eqs. \ref{eq:given_rotation_transformations} and \ref{eq:given_reflection_transformations}).

\begin{equation}
 G^{+}(\theta_{p, i}) = \begin{bmatrix} \cos(\theta_{p, i}) & -\sin(\theta_{p, i}) \\ \sin(\theta_{p, i}) & \cos(\theta_{p, i}) \end{bmatrix}
\label{eq:given_rotation_transformations}
\end{equation}

\begin{equation}
     G^{-}(\phi_{p, i}) = \begin{bmatrix} \cos(\phi_{p, i}) & \sin(\phi_{p, i}) \\ \sin(\phi_{p, i}) & -\cos(\phi_{p, i}) \end{bmatrix}
     \label{eq:given_reflection_transformations}
\end{equation}  

Then, the rotations and reflections are applied only to the subject embedding as describe in Eq. \ref{eq:rot_ref_subj_embedding}.
\begin{equation}
    q^{H}_{rot}=W^{rot}_{\Theta_{p}}e^{H}_{s} \qquad q^{H}_{ref}=W^{ref}_{\Phi_{p}}e^{H}_{s}
    \label{eq:rot_ref_subj_embedding}
\end{equation}
Furthermore, to represent complex relations that can be a mixture of rotation and reflection, \textsc{AttH} utilizes an hyperbolic attention mechanism. The attention scores $\alpha_{q^{H}_{rot}}^{p}$ and $\alpha_{q^{H}_{ref}}^{p}$ are computed in the tangent space by projecting the hyperbolic rotation embedding $q^{H}_{rot}$ and hyperbolic reflection embedding $q^{H}_{ref}$ with the logarithmic map (Eq. \ref{eq:logmap}). More specifically, \textsc{AttH} implements a tangent space average to implement the typical weighted average as proposed in \citet{liu-et-al-2019-hgn} and \citet{chami-et-al-2019-hgcn}. Then, the attention vector is mapped back to manifold using the exponential map (Eq. \ref{eq:expmap}). We have then:
\begin{equation}
\begin{split}
    ATT(q^{H}_{rot}, q^{H}_{ref}, p) = \: &\exp_\textbf{0}^{c_{p}}(\alpha_{q^{H}_{rot}}\log_\textbf{0}^{c_{p}}(q^{H}_{rot}) \\ &+ \alpha_{q^{H}_{ref}}\log_\textbf{0}^{c_{p}}(q^{H}_{ref}))
\end{split}
\label{eq:hyperbolic_attention_mechanism}
\end{equation}
\textsc{AttH} finally applies a translation of the hyperbolic relation embedding $r_{p}^{H}$ over the resulting attention vector (Eq. \ref{eq:hyperbolic_attention_mechanism}). As mentioned in \citet{chami-etal-2020-low}, translations help to move between different levels of the hierarchy. 
\begin{equation}
    Q(s, p) = ATT(q^{H}_{rot}, q^{H}_{ref}, p) \oplus_{c_{p}} r_{p}^{H}
\end{equation}
The scoring function is similar to the one used in \citet{balazevic-et-al-2019-murp} and \citet{han-etal-2020-dyernie} defined as: 
\begin{equation}
    s(s,p,o,t) = -d^{c_{p}}(Q(s,p), e_{o}^{H})^{2} + b_{s} + b_{o}
\end{equation}
where $b_{s}$ and $b_ {o}$ stand for the subject and object biases, respectively.

We then propose \textsc{Hercules}, a time-aware extension of \textsc{AttH}. \textsc{Hercules} redefines the curvature of the manifold as being the product of both relation and time as illustrated in Eq. \ref{eq:hercules_curvature}: 
\begin{equation}
    c_{p}^{t} = \sigma(\mu_{p} \times \tau_{t})
    \label{eq:hercules_curvature}
\end{equation}
with $\tau_{t}$ a learnable parameter $\in \mathbb{R}$.
A trade-off therefore exists between relation and time. A low value of either $\mu_{p}$ or $\tau_{t}$ will lead to a flatter space while higher values will tend to a more hyperbolic space. The main intuition of \textsc{Hercules} is that both relation and time directly adjust the geometry of the manifold such that the positions of projected entities are relation-and-time-dependent. This is advantageous in that no additional temporal parameters per entity are needed. Since the whole geometry has changed for specific relation and time, all future projections onto that manifold will be aligned to the corresponding relation and timestamp. We investigate different curvature definitions and time translation in our experiments (see Section \ref{section:experiments}). The scoring function of \textsc{Hercules} remains same as \textsc{AttH}. 

When learning hyperbolic parameters, the optimization requires to utilize a Riemannian gradient \cite{bonnabel-et-al-2011-rsgd}. However, proven to be challenging, we instead learn all embeddings in the Euclidean space. The embeddings can then be mapped to the manifold using the exponential map (Eq. \ref{eq:expmap}). This allows the use of standard Euclidean optimization strategies.

\begin{table*}[!ht]
\centering
\begin{tabular}{cc}
\hline \textbf{Model} & \textbf{Parameters} \\ \hline
\textsc{TTransE} & $(\lvert \mathcal{E} \rvert + 2\lvert \mathcal{R} \rvert + \lvert \mathcal{T} \rvert ) \cdot n $ \\
\textsc{TComplEx} & $(2\lvert \mathcal{E} \rvert + 4\lvert \mathcal{R} \rvert + 2\lvert \mathcal{T} \rvert ) \cdot n $ \\
\textsc{HyTe} & $(\lvert \mathcal{E} \rvert + 2\lvert \mathcal{R} \rvert + \lvert \mathcal{T} \rvert ) \cdot n $ \\
\textsc{ATISE} & $(\lvert \mathcal{E} \rvert + 2\lvert \mathcal{R} \rvert) \cdot n $  \\
\textsc{TERO} & $(\lvert \mathcal{E} \rvert + 2\lvert \mathcal{R} \rvert + \lvert \mathcal{T} \rvert ) \cdot n $ \\
\textsc{DyERNIE} & $2 \cdot (\lvert \mathcal{E} \rvert + 2\lvert \mathcal{R} \rvert) \cdot n + 2 \cdot \lvert \mathcal{E} \rvert $ \\
\textsc{AttH} & $(\lvert \mathcal{E} \rvert + 2\lvert \mathcal{R} \rvert) \cdot n + \lvert \mathcal{E} \rvert + 2\lvert \mathcal{R} \rvert (1+3n) $ \\
\textsc{Hercules} & $(\lvert \mathcal{E} \rvert + 2\lvert \mathcal{R} \rvert) \cdot n + \lvert \mathcal{E} \rvert + 2\lvert \mathcal{R} \rvert (1+3n) + \lvert \mathcal{T} \rvert $ \\
\hline
\end{tabular}
\caption{\label{table-number_parameters} Number of parameters per models with respect to the embedding dimension \textit{n}.}
\end{table*}

\section{Experiments}
\label{section:experiments}
We outline in this section the experiments and evaluation settings.
\subsection{Datasets}
\label{section:datasets}
For fair comparisons, we test our model on same benchamark datasets used in previous works, \textit{i.e.} ICEWS14 and ICEWS05-15. Both datasets were constructed by \citet{garcia-duran-etal-2018-learning} using the Integrated Crisis Early Warning System (ICEWS) dataset \cite{boschee-et-al-2018-icews}. ICEWS provides geopolitical information with their corresponding (event) date, \textit{e.g.} $\langle$\textit{Barack Obama, visits, France, 2009-03-11}$\rangle$. More specifically, ICEWS14 includes events that happened in 2014 whereas ICEWS05-15 encompasses facts that appeared between 2005 and 2015. We give the original datasets statistics in Table \ref{table:datasets_stats}. To increase the number of samples, for each quadruplet $\langle$\textit{s, p, o, t}$\rangle$ we add $\langle$\textit{s, $p^{-1}$, o, t}$\rangle$, where $p^{-1}$ is the inverse relation of \textit{p}. This is a standard data augmentation technique usually used in LP \cite{balazevic-et-al-2019-murp, goel-et-al-2020-diachronic, han-etal-2020-dyernie}.

\begin{table*}
    \centering
    \begin{tabular}{cccccccc}
    \hline
         \textbf{Datasets} & $\boldsymbol{\lvert \mathcal{E} \rvert}$ & $\boldsymbol{\lvert \mathcal{R} \rvert}$ & $\boldsymbol{\lvert \mathcal{T} \rvert}$ & \textbf{Training} & \textbf{Validation} & \textbf{Test}  \\ \hline
         ICEWS14 & 7,128 & 230 & 365 & 72,128 & 8,941 &
         8,963 \\ \hline
         ICEWS05-15 & 10,488 & 251 & 4017 & 368,962 & 46,275 & 46,092 \\
         \hline
 
 \end{tabular}
    \caption{ICEWS14 and ICEWS05-15 Datasets Statistics}
    \label{table:datasets_stats}
\end{table*}

\subsection{Evaluation Protocol \& Metrics}
\label{subsection:evaluation_protocol_metrics}
Given a (golden) test triple $\langle$\textit{s, p, o, t}$\rangle$, for each entity $s' \in \mathcal{E}$, we interchange the subject \textit{s} with $s'$ and apply the scoring function $f$ on the resulting query $\langle s'$\textit{, p, o, t}$\rangle$. Since replacing \textit{s} by all possible entity $s'$ may end up with a correct facts, we filter out those valid quadruplets and give them extremely low scores to avoid correct quadruplets to be scored higher than the tested quadruplet in final ranking \cite{bordes-et-al-2013-transe}. We then rank the entities based on their scores in descending order. We store the rank of the correct entity \textit{s} noted $z_{s}$. Thus, the model should maximize the returned score for the entity \textit{s} such that $z_{s}=1$. The same process is done using the object $o$.

To evaluate our models, we make use of the Mean Reciprocal Rank (MRR). We also provide the Hits@1 (H@1), Hits@3 (H@3) and Hits@10 (H@10) which assess on the frequency that the valid entity is in the top-1, top-3 and top-10 position, respectively.

\subsection{Implementation Details}
To ensure unbiased comparability, the same training procedure and hyper-parameters are shared for \textsc{AttH} and \textsc{Hercules}.\footnote{We used the official implementation of \textsc{AttH} available at \url{https://github.com/HazyResearch/KGEmb}. We adapted it to implement \textsc{Hercules}.} Number of epochs and batch size were set to 500 and 256, respectively. We minimized the cross-entropy loss using negative sampling, where negative samples corrupt the valid object only (uniformly selected). After validation, we noticed that best results were obtained using 500 negative samples. We chose the Adam optimizer \cite{kingma-et-al-2014-adam} with an initial learning rate of 0.001. The final models for evaluation were selected upon the MRR metric on the validation set. We re-train \textsc{ATISE} and \textsc{TeRo} using the same parameters as mentionned in \citet{xu-et-al-2019-atise} and \citet{xu-et-al-2020-tero} but varying dimensions.\footnote{We used the official implementation available at \url{https://github.com/soledad921/ATISE}}


\subsection{Results}
We report models performances on the link prediction task on TKGs. Additional analyses on the results are also given.
\subsubsection{Link Prediction results}
We provide link prediction results on ICEWS14 and ICEWS05-15 for \textsc{AttH}, \textsc{Hercules} and different models from the literature. As \citet{han-etal-2020-dyernie}, we adopted a dimension analysis to investigate behaviors and robustness of approaches. When possible, we re-run official implementation of models. Otherwise, official or best results in literature are reported. Results are shown in Table \ref{table:results_link_prediction}.

As expected, hyperbolic-based strategies (\textit{i.e.} \textsc{DyERNIE}, \textsc{AttH} and \textsc{Hercules}) perform much better at lower dimensions, outperforming most of other approaches with ten times less dimensions. We report an average absolute gain of 11.6\% points in MRR with only 10 dimensions over the median performance of other approaches with 100 dimensions. This strengthens the effectiveness of hyperbolic geometry to induce high-quality embeddings with few parameters.

Astonishingly, we notice that \textsc{AttH} model is highly competitive despite the absence of time parameter. \textsc{AttH} exhibits new state-of-the-art or statistically equivalent performances compared to \textsc{DyERNIE} and \textsc{Hercules}. We remark no statistically significant differences in performances between hyperbolic models.\footnote{\label{stests}We performed the Mixed-Factorial Analysis of Variance (ANOVA), in which the independent variables are \textit{the dimension} and \textit{the model} and the dependent variable is \textit{the metric}. We consider two groups one for each dataset. We report \textit{p}-values of 0.842, 0.872, 0.926 and 0.229 for MRR, H\char64{}1, H\char64{}3 and {H\char64{}10} respectively.} Importantly, unlike other research carried out in this area, time information here does not lead to any notable gain. This seems to indicate that other parameters should be considered. We examine this phenomenon in section
\ref{section:time_is_all_you_need_negative}.

On ICEWS14, for \textit{dim} $\in \{20, 40, 100\}$, both \textsc{AttH} and \textsc{Hercules} outperform \textsc{DyERNIE} by a large margin. We witness an improvement of 2.5\% and 5\% points in MRR and Hits@1 with 100-dimensional embeddings. On ICEWS05-15, \textsc{Atth} and \textsc{Hercules} yield comparable achievements with the state-of-the-art. In contrast to \textsc{DyERNIE}, it is noteworthy that \textsc{AttH} and \textsc{Hercules} utilize a single manifold while reaching top performances. 

We also distinguish tempered results on Hits@10 metric for \textsc{AttH} and \textsc{Hercules} models. This suggests that during optimization, \textsc{AttH} and \textsc{Hercules} favor ranking some entities on top while harming the representation of others.



\begin{table*}[th!]
\centering
\begin{tabular}{c|c|cccc|cccc}
\hline
\multicolumn{2}{c}{Datasets} & \multicolumn{4}{|c}{{ICEWS14 (filtered)}} & \multicolumn{4}{|c}{{ICEWS05-15 (filtered)}} \\
\hline
{\textit{dim}} & {Model} & {MRR} & {H@1} & {H@3} & {H@10} & {MRR} & {H@1} & {H@3} & {H@10} \\ \hline 


{} & \textsc{ATISE\textsuperscript{\ding{61}}} & {18.0} & {3.03} & {23.9} & {48.7} & {15.9} & {4.35} & {19.22} & {41.0} \\

{} & \textsc{TeRo\textsuperscript{\ding{61}}} & {7.25} & {2.39} & {6.40} & {16.6} & {10.3} & {3.54} & {10.1} & {23.2} \\

{10} & \textsc{DyERNIE\textsuperscript{\ding{83}}} & \textbf{46.2} & \textbf{36.0} & {51.1} & \underline{66.3} & \textbf{58.9} & \textbf{50.5} & \textbf{63.2} & \textbf{75.1} \\
\cline{2-10}

{} & \textsc{Hercules} & \underline{46.0} & \underline{34.9} & \textbf{52.4} & {66.0} & \underline{54.7} & \underline{43.8} & \underline{61.8} & {73.2} \\

{} & \textsc{AttH} & {45.6} & {34.2} & \underline{52.0} & \textbf{66.4} & {49.9} & {34.4} & {61.6} & \underline{73.6} \\

\hline 

{} & \textsc{ATISE\textsuperscript{\ding{61}}} & {19.1} & {1.28} & {28.2} & {54.7} & {24.5} & {7.67} & {32.3} & {59.2} \\

{} & \textsc{TeRo\textsuperscript{\ding{61}}} & {24.5} & {13.8} & {28.01} & {46.3} & {27.1} & {13.5} & {33.3} & {54.1} \\

{20} & \textsc{DyERNIE\textsuperscript{\ding{83}}} & {53.9} & {44.2} & {58.9} & \textbf{72.7} & \textbf{64.2} & \textbf{56.5} & \textbf{68.2} & \textbf{79.0} \\

\cline{2-10}

{} & \textsc{Hercules} & \textbf{55.5} & \textbf{47.2} & \underline{59.4} & \underline{71.4} & {63.2} & {55.2} & \underline{67.7} & \underline{77.6} \\

{} & \textsc{AttH} & \underline{55.2} & \underline{46.7} & \textbf{59.7} & \underline{71.4} & \underline{63.5} & \underline{55.8} & \underline{67.7} & {77.5} \\

\hline 

{} & \textsc{ATISE\textsuperscript{\ding{61}}} & {38.4} & {23.3} & {47.6} & {67.3} & {35.7} & {19.2} & {44.3} & {69.1} \\

{} & \textsc{TeRo\textsuperscript{\ding{61}}} & {35.1} & {22.7} & {40.5} & {60.8} & {28.3} & {12.7} & {35.3} & {60.5} \\

{40} & \textsc{DyERNIE\textsuperscript{\ding{83}}} & {58.8} & {49.8} & {63.8} & \textbf{76.1} & \textbf{68.9} & {61.8} & \textbf{72.8} & \textbf{82.5} \\

\cline{2-10}

{} & \textsc{Hercules} & \underline{61.2} & \underline{54.3} & \underline{64.7} & {74.1} & \underline{68.5} & \textbf{62.1} & \underline{72.0} & \underline{80.9} \\

{} & \textsc{AttH} & \textbf{61.7} & \textbf{54.5} & \textbf{65.4} & \underline{75.4} & \underline{68.5} & \underline{62.0} & {71.9} & {80.6} \\

\hline 

{} & \textsc{TransE\textsuperscript{\ding{83}}} & {30.0} & {14.8} & {42.7} & {60.1} & {30.4} & {13.3} & {42.4} & {61.1} \\ 

{} & \textsc{DistMult\textsuperscript{\ding{83}}} & {57.5} & {46.9} & {64.2} & {77.9} & {47.1} & {33.6} & {55.1} & {72.5} \\ 

{} & \textsc{ComplEx\textsuperscript{\ding{83}}} & {49.3} & {36.6} & {56.2} & {74.2} & {39.0} & {22.9} & {49.2} & {68.4} \\ 

{} & \textsc{TTransE\textsuperscript{\ding{83}}} & {34.4} & {25.7} & {38.3} & {51.3} & {35.6} & {15.4} & {51.1} & {67.6} \\ 

{} & \textsc{TComplEx\textsuperscript{\ding{83}}} & {31.8} & {12.9} & {45.7} & {63.0} & {45.1} & {36.3} & {49.2} & {62.0} \\

{100} & \textsc{HyTE\textsuperscript{\ding{83}}} & {33.1} & {6.8} & {54.5} & {73.6} & {38.1} & {7.6} & {65.0} & {80.4} \\

{} & \textsc{ATISE\textsuperscript{\ding{61}}} & {52.2} & {41.0} & {60.0} & {72.7} & {47.0} & {32.4} & {55.5} & {76.4} \\

{} & \textsc{TeRo\textsuperscript{\ding{61}}} & {45.4} & {34.0} & {52.2} & {67.0} & {41.1} & {26.3} & {48.9} & {71.7} \\

{} & \textsc{DyERNIE\textsuperscript{\ding{83}}} & {66.9} & \underline{59.9} & \underline{71.4} & \textbf{79.7} & \textbf{73.9} & \underline{67.9} & \textbf{77.3} & \textbf{85.5} \\

\cline{2-10}

{} & \textsc{Hercules} & \underline{69.4} & \textbf{65.0} & \underline{71.4} & {77.9} & {73.5} & \textbf{68.6} & \underline{76.1} & \underline{82.9} \\

{} & \textsc{AttH} & \textbf{69.5} & \textbf{65.0} & \textbf{71.5} & \underline{78.2} & \underline{73.6} & \textbf{68.6} & {76.0} & \underline{82.9} \\

\hline 

\end{tabular}
\caption{Link prediction results on ICEWS14 and ICEWS05-15 datasets: (\ding{61}) results are obtained using the official implementation of \citet{xu-etal-2020-tero}, (\ding{83}) results are taken from \citet{han-etal-2020-dyernie}. For each dimension (\textit{i.e. dim}), best results are in bold and second-to-best underlined. No statistically significant differences in performance are observed between \textsc{DyERNIE}, \textsc{Hercules} and \textsc{AttH}.\footnoteref{stests}}
\label{table:results_link_prediction}
\end{table*}

\subsubsection{Is Time \textit{All} You Need ?}
\label{section:time_is_all_you_need_negative}
In this section, we investigate the influence of the temporal parameter on performances.

First, besides time translation, we probe different curvature definitions to identify fluctuation in performances. We analyze how time information alters the LP results by adding time as part of the curvature (\textit{i.e.} \textsc{Hercules}) and as a translation. We also explore if incorporating the Euclidean dot product of the subject and object embeddings (noted $\langle e_{s}^{\mathbb{E}},\:e_{o}^{\mathbb{E}}\rangle$) into the curvature helps to learn a better geometry. An ablation study is given in Table \ref{tab:ablation}.
 
\begin{table*}[ht!]
    \centering
    \begin{tabular}{cccc||c|c|c|c}
    \hline
    \begin{tabular}[c]{@{}c@{}}Relation\\ Curvature\end{tabular} & \begin{tabular}[c]{@{}c@{}}Time\\ Curvature\end{tabular} & \begin{tabular}[c]{@{}c@{}}Time\\ Translation\end{tabular} & \begin{tabular}[c]{@{}c@{}}$\langle e_{s}^{\mathbb{E}},\:e_{o}^{\mathbb{E}}\rangle$\\ Curvature\end{tabular} & {MRR} & {H@1} & {H@3} & {H@10} \\ \hline
    {\ding{51}} &\ding{55}&\ding{55}&\ding{55}& {61.7} & {54.5} & {65.4} & {75.4} \\ 
    {\ding{51}} &{\ding{51}}&\ding{55}&\ding{55}& {61.2} & {54.3} & {64.7} & {74.1} \\ 
    {\ding{51}} &{\ding{51}}&{\ding{51}}&\ding{55}& {60.1} & {52.1} & {64.5} & {75.0} \\ 
    {\ding{51}} &{\ding{51}}&{\ding{51}}&{\ding{51}}& {49.5} & {38.9} & {55.4} & {69.2} \\ \hline
    \end{tabular}
    \caption{Ablation study: Link prediction results on ICEWS14 using \textsc{AttH} (\textit{dim} = 40) with different curvature definitions and time translation applied.}
    \label{tab:ablation}
\end{table*}

Albeit counter-intuitive, we observe that our results corroborate with our initial finding: time information is not the culprit of our high performances. More strikingly, a simple relational curvature 
(\textit{i.e.}\ \textsc{AttH}) is sufficient to perform best on ICEWS14 (\textit{dim} = 40). Neither the inclusion of a time translation, similarly to \textsc{TTransE}, nor the Euclidean dot product provide interesting outcomes.

\begin{figure}[th]
    \centering
    \includegraphics[width=\linewidth]{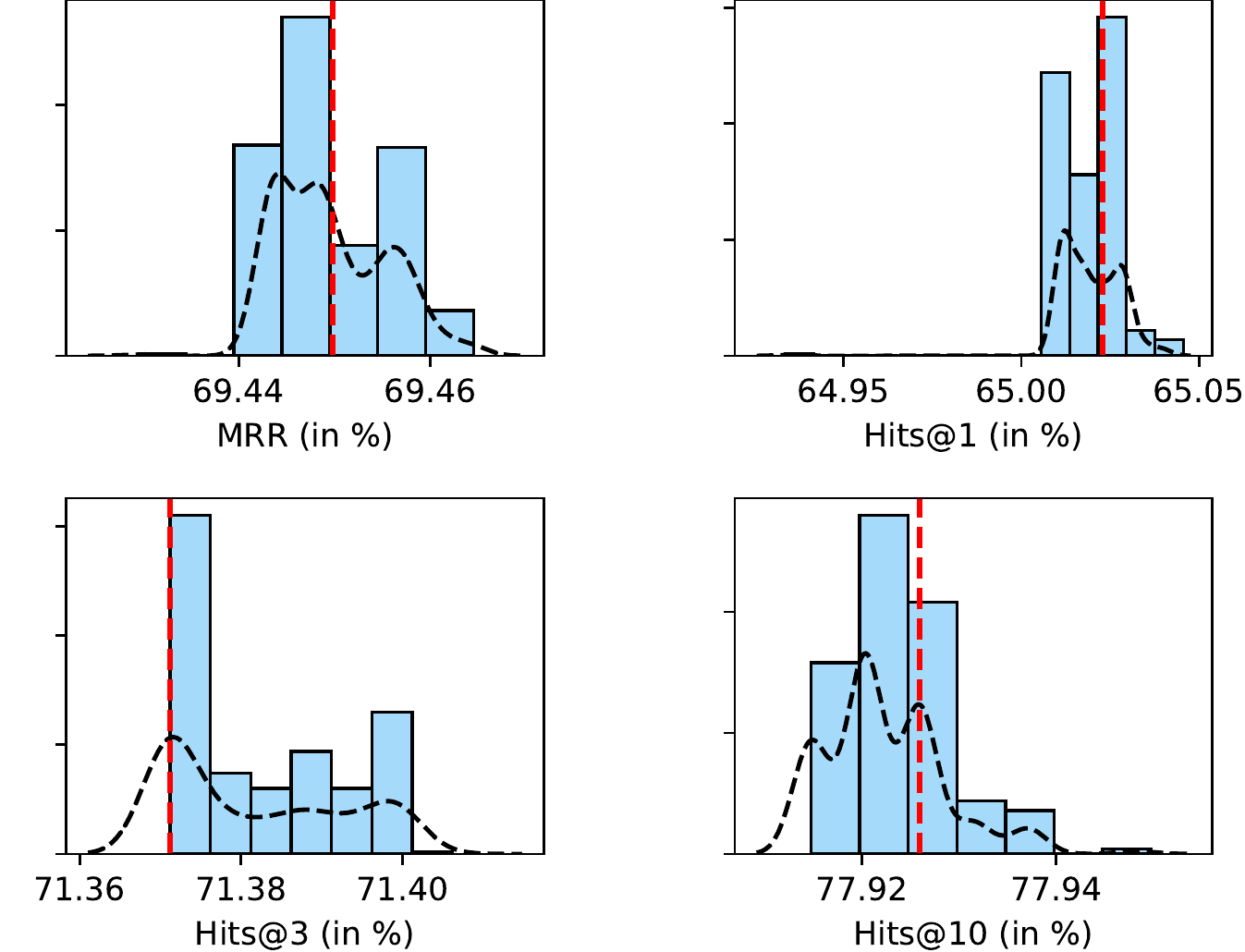}
    \caption{Distribution of performances of \textsc{Hercules} on temporally-corrupted quadruplets from ICEWS14 with \textit{dim}=100. A smooth approximation of the distribution is drawn as a dashed black curve. The reference performance is indicated with a dashed red line.}
    \label{fig:icews14_hercules_probing}
\end{figure}

We then probe the sensitivity of \textsc{Hercules} towards temporal feature by performing LP with incorrect timestamps. Our intuition is to inspect whether feeding invalid timestamps during evaluation exhibits significant variation or not compared to the reference performances, \textit{i.e.} LP results with initial (non-corrupted) testing samples. To do so, for each testing quadruplet, we replace the (correct) time parameter with each possible timestamp from $\mathcal{T}$. We therefore collect multiple LP performances of \textsc{Hercules} corresponding to each distinct timestamp. We plot the distribution of resulting performances of temporally-corrupted quadruplets from ICEWS14 (\textit{dim}=100) in Fig. \ref{fig:icews14_hercules_probing}. We can observe that despite erroneous timestamps, LP results show insignificant discrepancies with the initial \textsc{Hercules} performance (dashed red line). The standard deviations from \textsc{Hercules} reference performance for MRR, H@1, H@3, H@10 metrics are $1.78 \times 10^{-4}$, $3.18 \times 10^{-3}$, $1.05 \times 10^{-2}$, $3.55 \times 10^{-3}$ respectively. This indicates that \textsc{Hercules} gives little importance to the time parameter and thus only relies on the entity and the predicate to perform knowledge graph completion. This further highlights our finding that timestamp is not responsible for our attracting performances.

\begin{figure}[ht!]
    \centering
    \includegraphics[width=\linewidth]{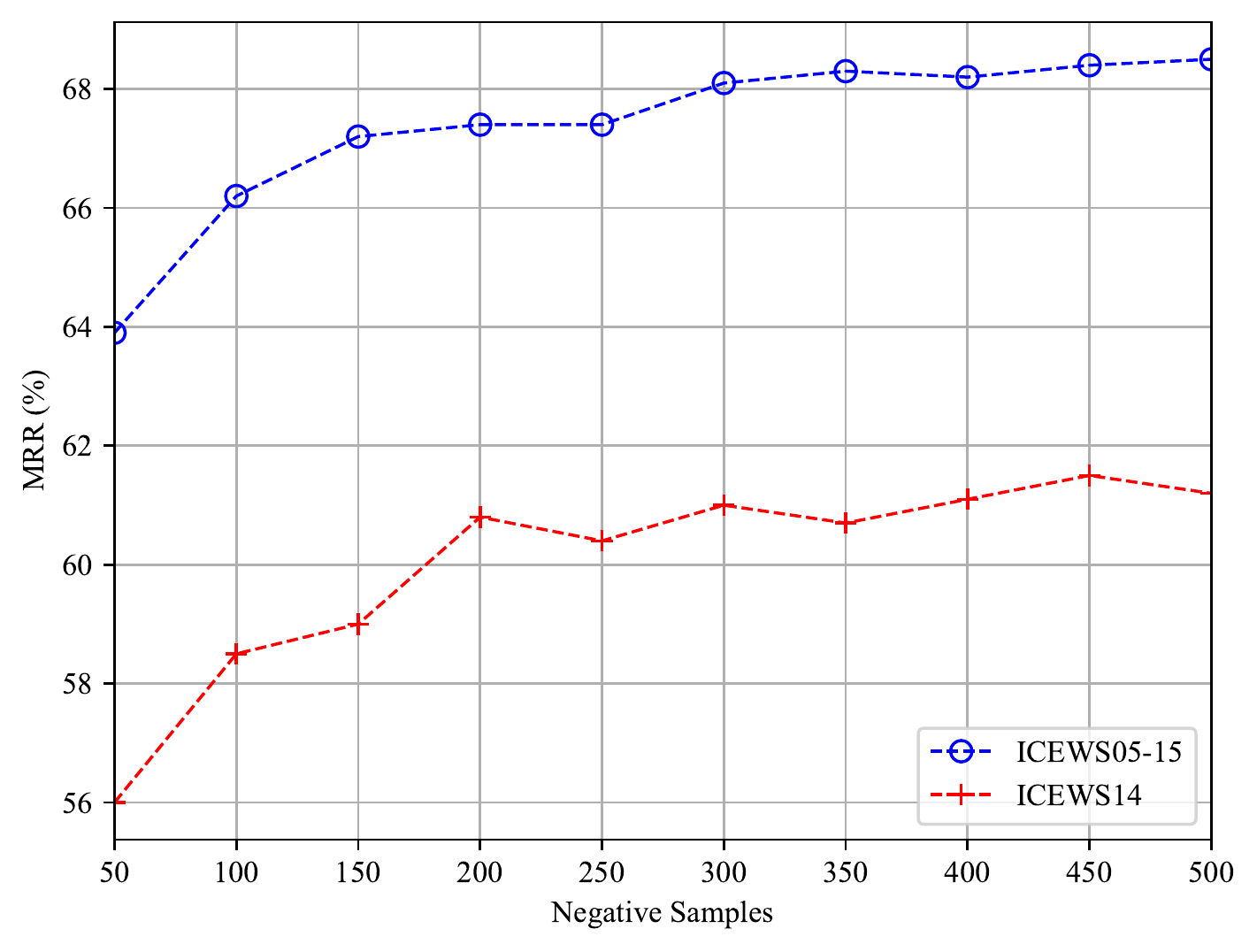}
    \caption{Performance of \textsc{Hercules} (\textit{dim} = 40) on ICEWS14 and ICEWS05-15 with varying number of negative samples.}
    \label{fig:neg_sampling}
\end{figure}

We therefore assume that the optimization procedure may be involved. We consequently question the effect of negative sampling. Precisely, we train \textsc{Hercules} with \textit{dim} = 40 by tuning the number of negative samples between 50 to 500. We plot the learning curves in Fig. \ref{fig:neg_sampling}.

For both, ICEWS14 and ICEWS05-15, negative sampling shows considerable gain as the number of samples increases. We record an absolute gain of 5\% points in MRR from 50 to 500 samples. We can see a rapid growth in MRR when the number of samples is inferior to 200. Adding 50 samples is equivalent to about 2\% points gain in MRR. Then, performances reach a plateau around 300 negative samples. We conjecture that a diversity in negative samples is enough to learn good representations. 
Notwithstanding that a large number of negative samples heavily constraints the location of entities in space, the resulting embeddings might benefit from it to be better positioned relatively to others.

We conclude that despite the present time parameter, an optimal negative sampling enables to reach new state-of-the-art outcome. Therefore, we argue that time is not the only parameter that should be considered when performing LP. We highlight that one should be raising awareness when training TKG representations to identify if time is truly helping to boost performances. 

\subsubsection{\textsc{AttH} versus \textsc{Hercules}}
\begin{figure}[th]
    \centering
    \includegraphics[width=\linewidth]{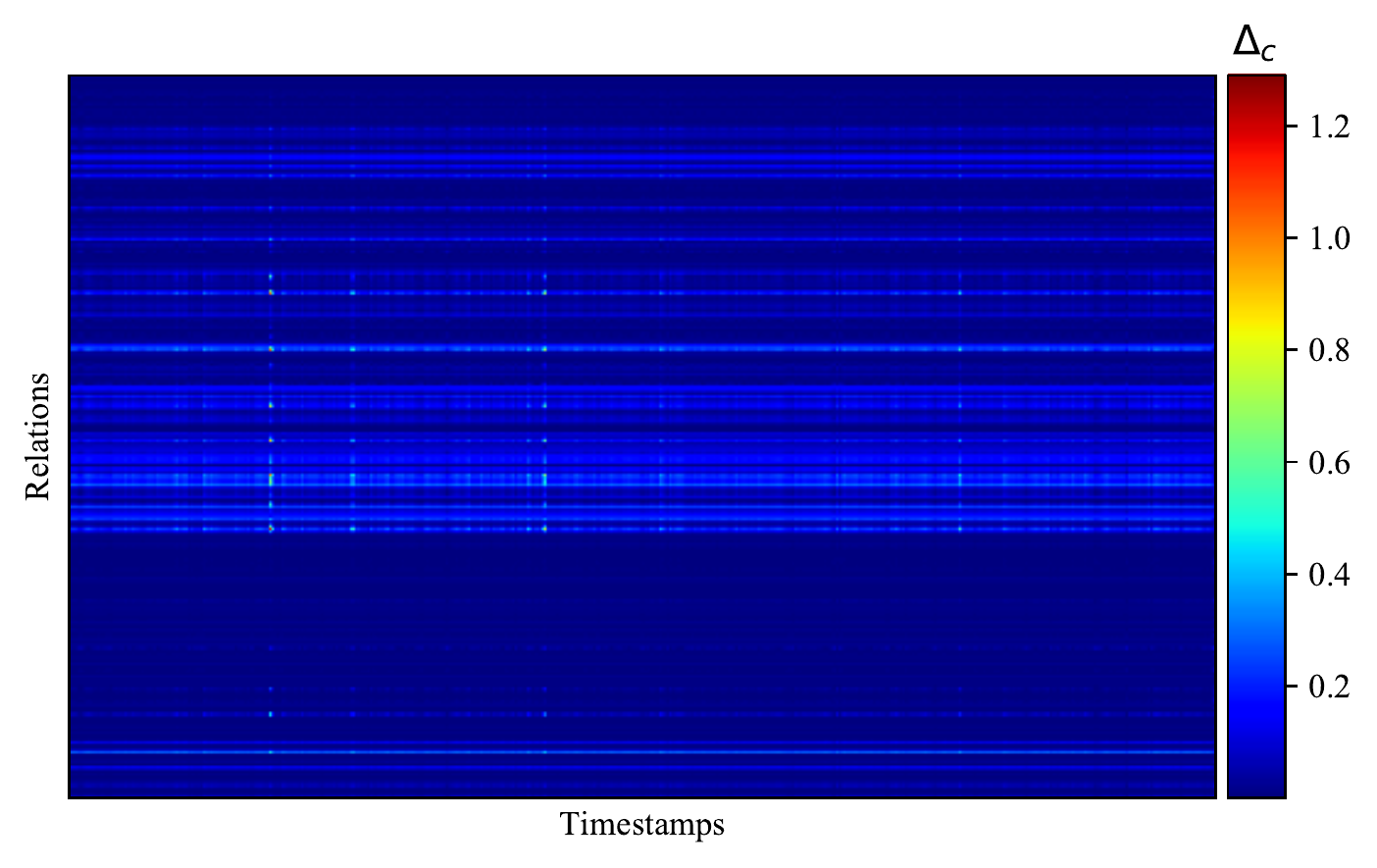}
    \caption{Absolute curvatures differences between \textsc{AttH} and \textsc{Hercules} given specific relation and time on ICEWS14 dataset (\textit{dim} = 40).}
    \label{fig:delta_c_icews14_atth_hercules_dim40}
\end{figure}

We explore here how the geometry of \textsc{Hercules} differs from \textsc{AttH}. To do so, we inspect the absolute difference of their learned curvatures $\Delta_{c}$. We plot $\Delta_{c}$ with respect to the relations and timestamps for \textit{dim} = 40 on ICEWS14 in Fig. \ref{fig:delta_c_icews14_atth_hercules_dim40}.\footnote{For readability, we don't plot $\Delta_{c}$ for reverse relations $p^{-1}$ (see Section \ref{section:datasets}).} Similar plots are given for ICEWS05-15 in Appendix \ref{appendix:atth_vs_hercules}.

\begin{figure}[th]
    \centering
    \includegraphics[width=\linewidth]{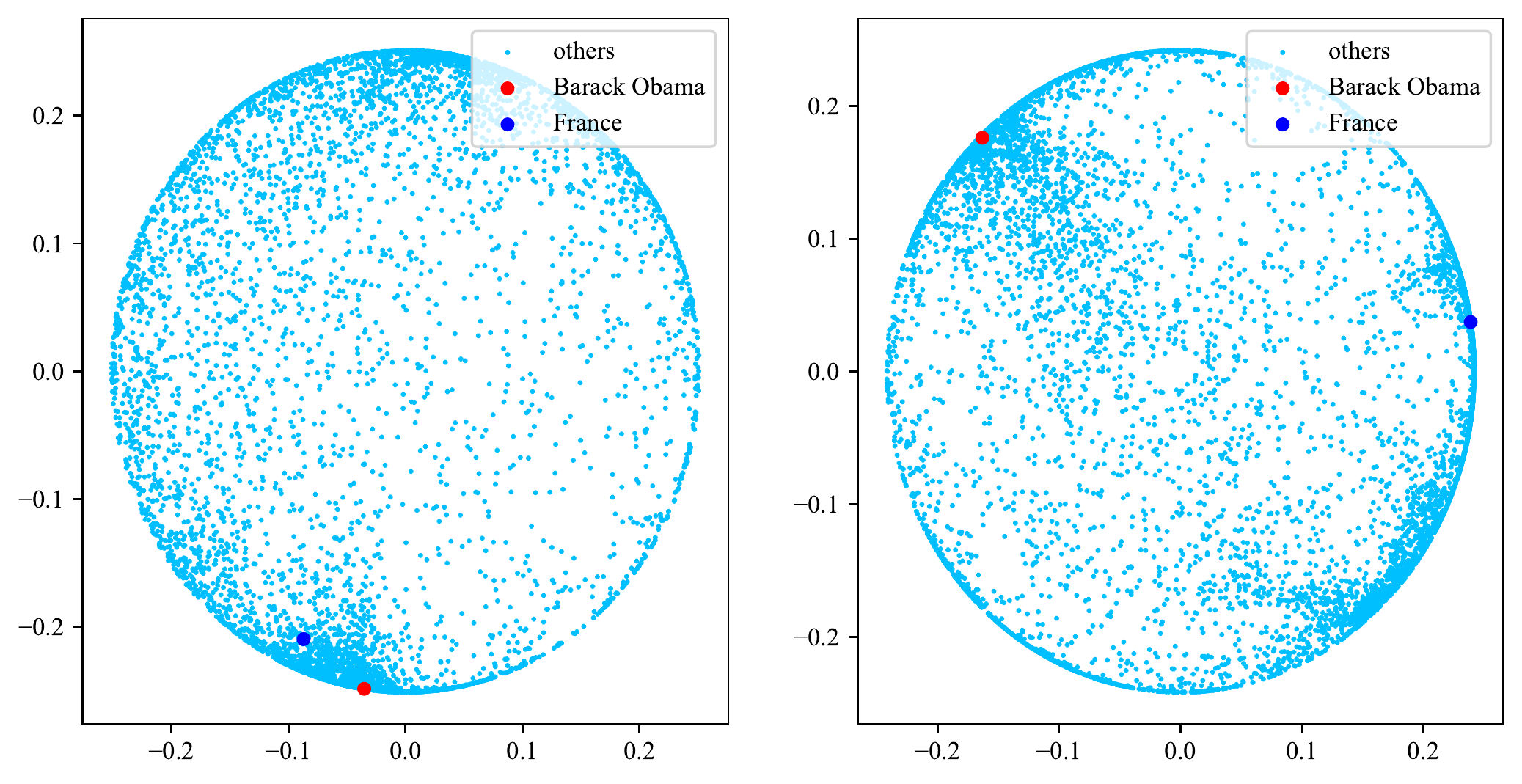}
    \caption{Illustration of two-dimensional hyperbolic entity embeddings learned by \textsc{AttH} (left) and \textsc{Hercules} (right) on ICEWS14 for predicate \lq \textit{make a visit}\rq and timestamp set to 01-01-2014.}
    \label{fig:2d_hyperbolic_embeddings_atth_hercules_icews14}
\end{figure}

Besides rare steep discrepancy in curvatures (\textit{i.e.} $\Delta_{c}$ $>$ 1.0), \textsc{Hercules} is akin to \textsc{AttH} concerning learned geometries. We report that 85.0\% and 95.6\% of $\Delta_{c}$'s are smaller than 0.1 on ICEWS14 and ICEWS05-15 respectively. We point out that some timestamps affect \textit{globally} all relations, albeit very limited. This can be seen in Fig. \ref{fig:2d_hyperbolic_embeddings_atth_hercules_icews14} by the aligned vertical strips. It indicates that \textsc{Hercules} uses its additional time parameter to learn a slightly different manifold but nonetheless quite similar to \textsc{AttH}.\footnote{Similar geometry $\neq$ Similar embeddings} We provide further analysis on the shifts of the curvatures while increasing embeddings dimension in Appendices \ref{appendix:atth_vs_atth} and \ref{appendix:hercules_vs_hercules}.

We depict an example of learned hyperbolic representation of ICEWS14 entities of \textsc{AttH} and \textsc{Hercules} for the relation \lq \textit{make a visit}\rq and timestamp set to 01-01-2014 in Fig. \ref{fig:2d_hyperbolic_embeddings_atth_hercules_icews14}. We plot embeddings of ICEWS05-15 in Appendix \ref{appendix:2d_hyperbolic_embeddings_atth_icews15}.

\section{Conclusion}
In this paper, we have demonstrated that without adding neither time information nor supplementary parameters, the \textsc{AttH} model astonishingly achieves similar or new state-of-the-art performances on link prediction upon temporal knowledge graphs. In spite of the inclusion of time with our proposed time-aware model \textsc{Hercules}, we have shown that negative sampling is sufficient to learn a good underlying geometry. In the future, we plan to explore new mechanisms to incorporate temporal information to improve performances of \textsc{AttH}.

\bibliographystyle{acl_natbib}
\bibliography{anthology,acl2021}

\label{sec:appendix}







\appendix

\section{Appendices}
\subsection{\textsc{AttH} versus \textsc{Hercules}}
\label{appendix:atth_vs_hercules}
\begin{figure}[ht]
    \centering
    \includegraphics[width=\linewidth]{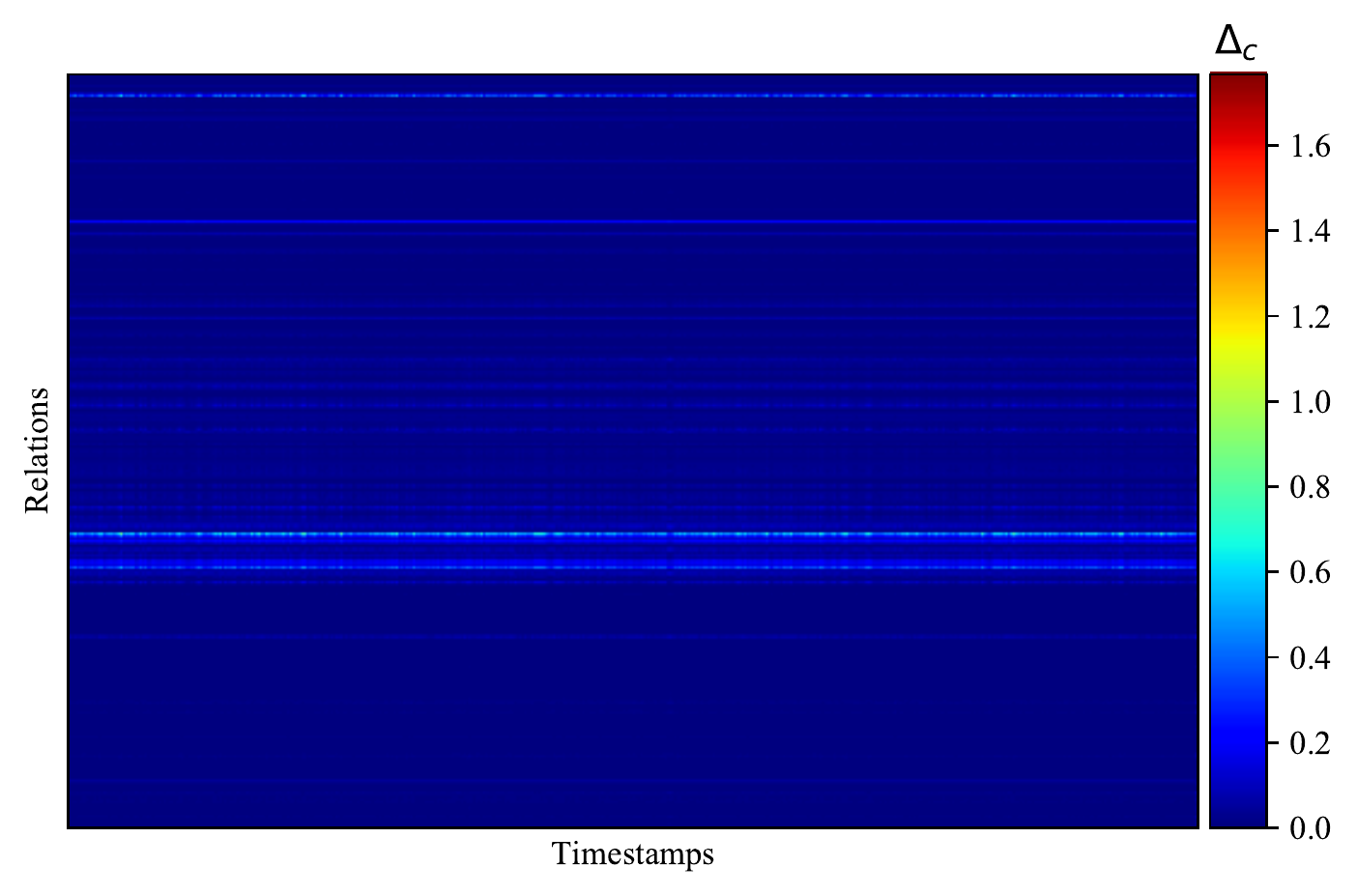}
    \caption{Absolute curvatures differences between \textsc{AttH} and \textsc{Hercules} given specific relation and time on ICEWS05-15 dataset (\textit{dim} = 40).}
    \label{fig:delta_c_icews05_atth_hercules_dim40}
\end{figure}

We plot the absolute difference in curvatures between \textsc{AttH} and \textsc{Hercules} for \textit{dim} = 40 on ICEWS05-15 in Fig. \ref{fig:delta_c_icews05_atth_hercules_dim40}.
As in ICEWS14 dataset, we observe that learned geometries on ICEWS15 dataset by \textsc{AttH} and \textsc{Hercules} are alike. Time is showing insubstantial impact on the curvature. 

    

\subsection{\textsc{AttH} versus \textsc{AttH}}
\label{appendix:atth_vs_atth}
\begin{figure}[htbp]
    \centering
    \includegraphics[width=\linewidth]{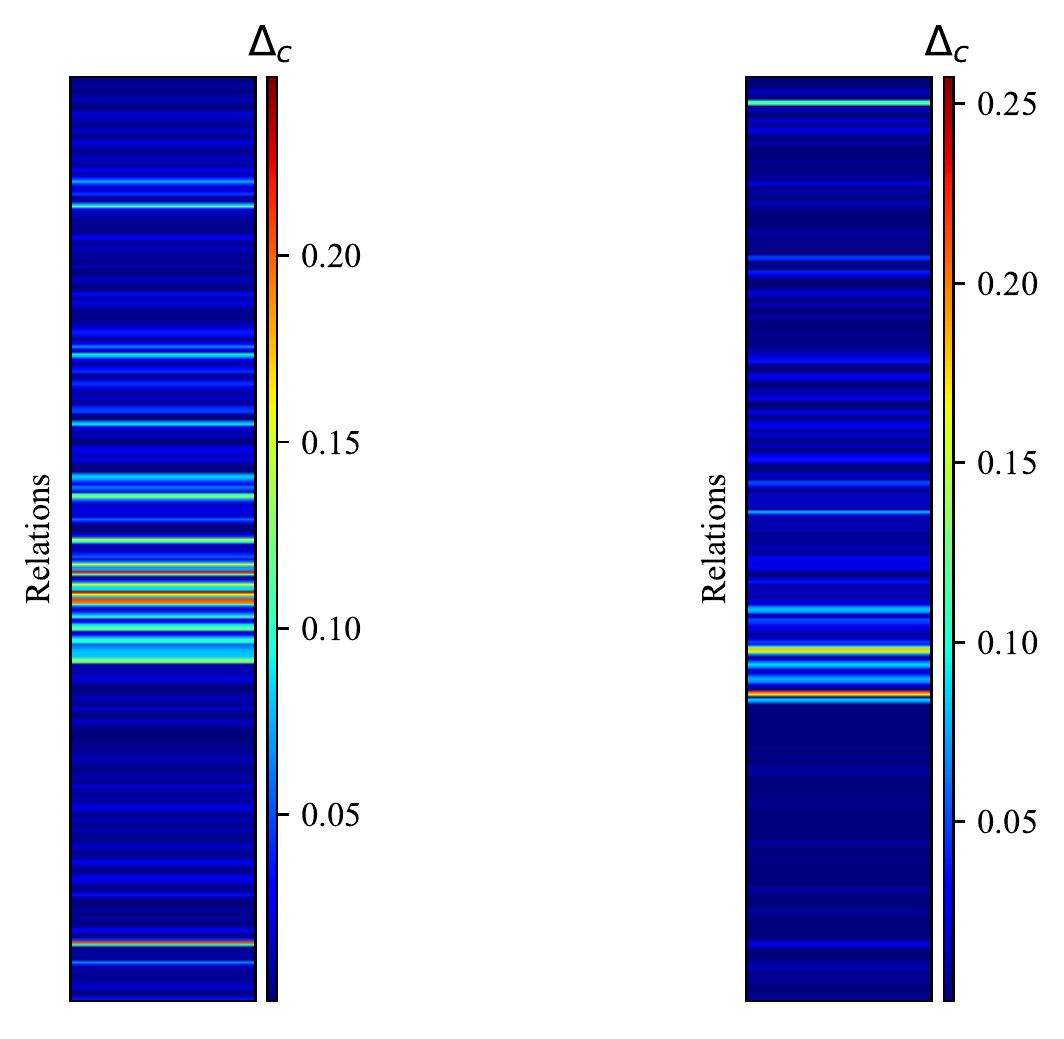}
    \caption{Absolute curvatures differences of \textsc{AttH} between \textit{dim} = 40 and \textit{dim} = 100 given specific relation and time on ICEWS14 dataset (left) and ICEWS05-15 dataset (right). No timestamps since \textsc{AttH} is a time-unaware model.}
    \label{fig:delta_c_icews14_icews05_dim40_atth_dim100_atth}
\end{figure}

We inspect how the curvature of \textsc{AttH} fluctuates while increasing the embedding dimension. We compare curvatures shifts between \textit{dim} = 40 and \textit{dim} = 100.
We plot $\Delta_{c}$ in Fig. \ref{fig:delta_c_icews14_icews05_dim40_atth_dim100_atth}.

The fluctuation in curvature is moderate while dimension of embeddings increases. We can see that variation is almost inferior to 0.2. This underlines that geometry does not change much for a specific relation when the dimension grows. For some relations, we note however that curvature is knowing significant dissimilitudes.

\subsection{\textsc{Hercules} versus \textsc{Hercules}}
\label{appendix:hercules_vs_hercules}
\begin{figure}[htbp]
    \centering
    \includegraphics[width=\linewidth]{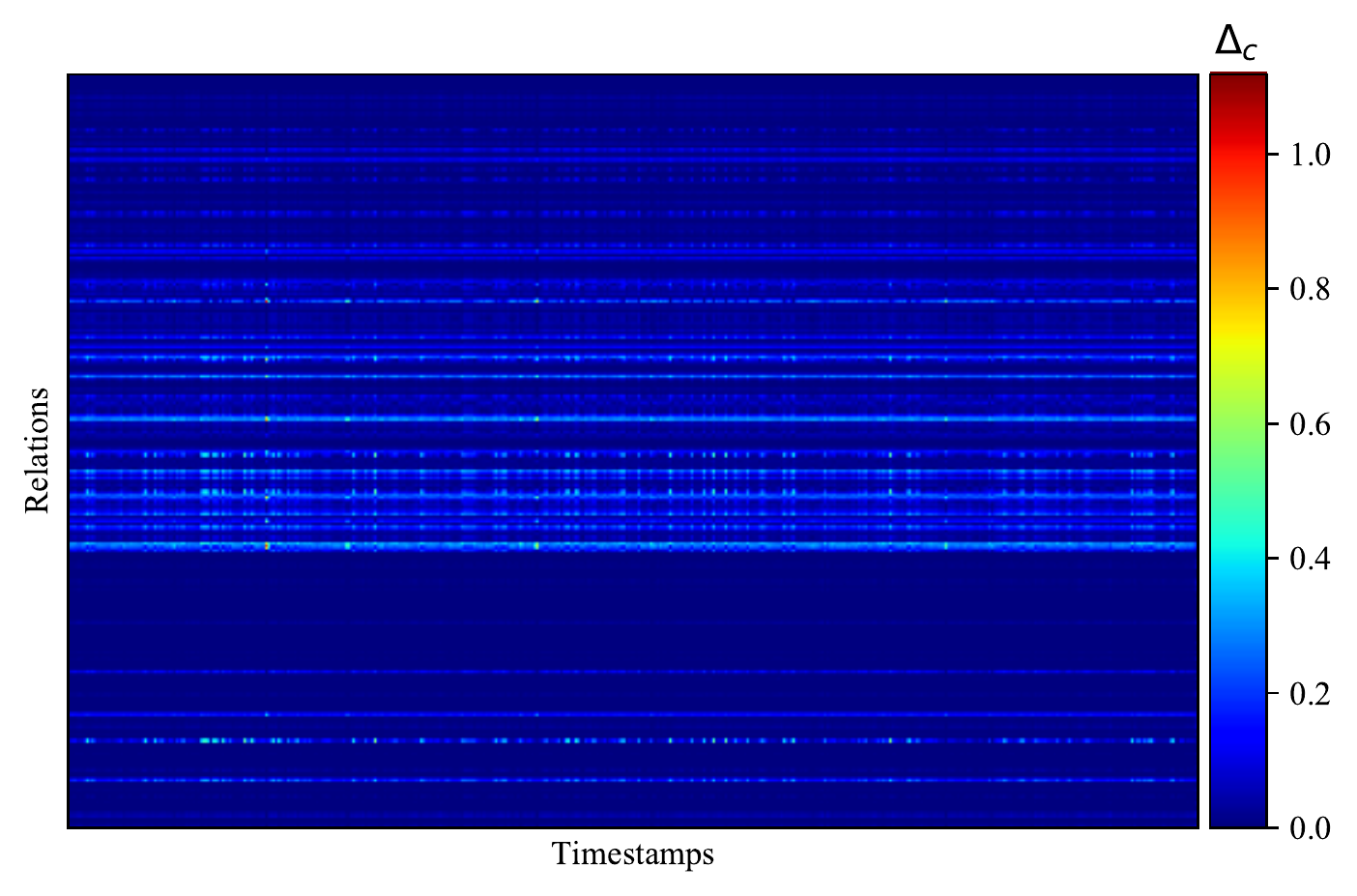}
    \caption{Absolute curvatures differences of \textsc{Hercules} between \textit{dim} = 40 and \textit{dim} = 100 given specific relation and time on ICEWS14 dataset.}
    \label{fig:delta_c_icews14_dim40_hercules_dim100_hercules}
\end{figure}

\begin{figure}[htbp]
    \centering
    \includegraphics[width=\linewidth]{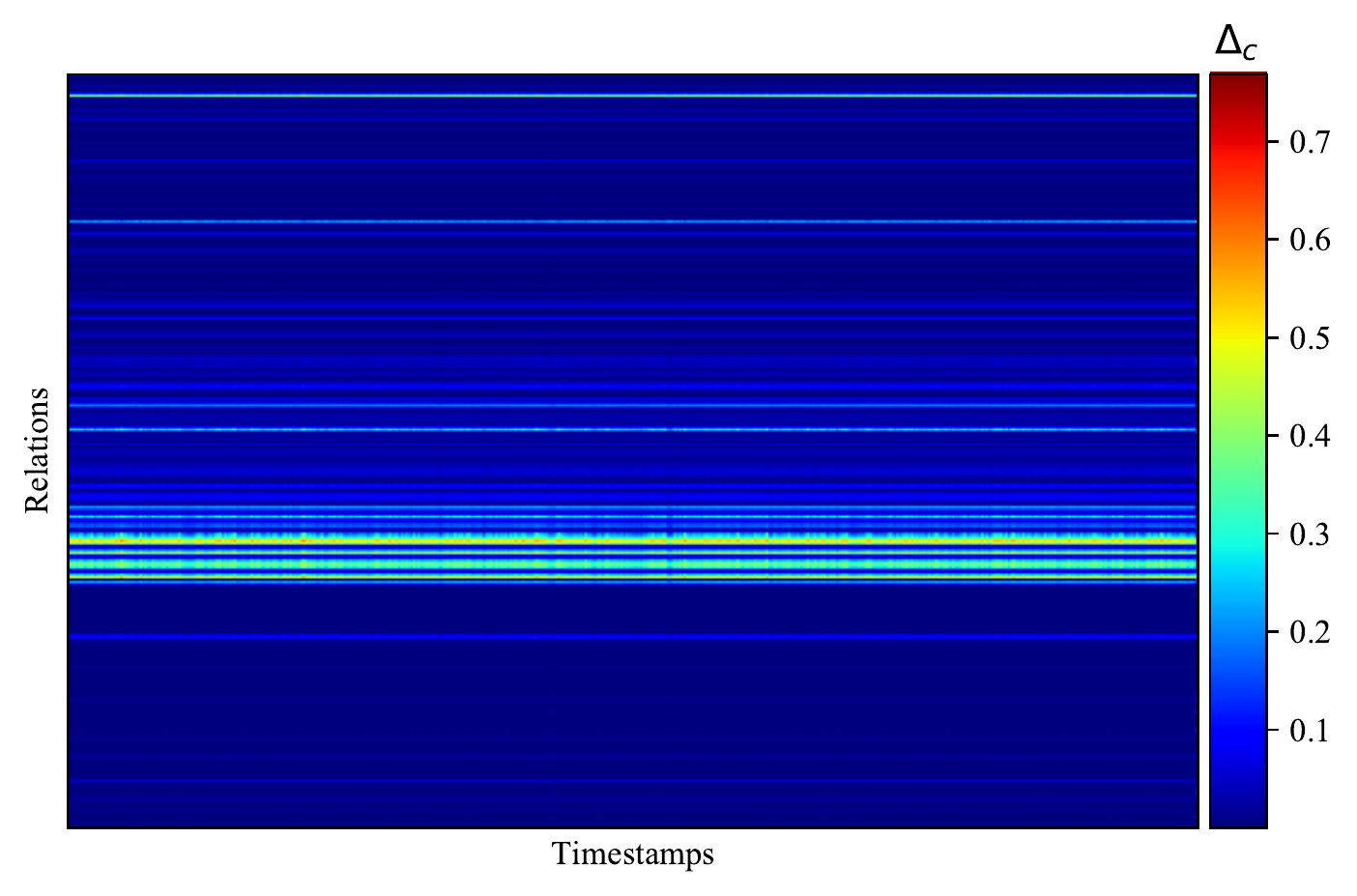}
    \caption{Absolute curvatures differences of \textsc{Hercules} between \textit{dim} = 40 and \textit{dim} = 100 given specific relation and time on ICEWS05-15 dataset.}
    \label{fig:delta_c_icews05_dim40_hercules_dim100_hercules}
\end{figure}

We report the dissimilarities in curvature for different relations and timestamps on ICEWS14 and ICEWS05-15 in Fig. \ref{fig:delta_c_icews14_dim40_hercules_dim100_hercules} and Fig. \ref{fig:delta_c_icews05_dim40_hercules_dim100_hercules}. 

We note that around 87.7\% and 91.1\% of variations are smaller than 0.1 on ICEWS14 and ICEWS05-15 respectively. This seems to indicate that despite the dimensionality gap, the learned geometry for each relation does not differ much between \textit{dim} = 40 and \textit{dim} = 100. 


\subsection{Two-Dimensional Hyperbolic Embeddings}
\label{appendix:2d_hyperbolic_embeddings_atth_icews15}
\begin{figure}[htbp]
    \centering
    \includegraphics[width=\linewidth]{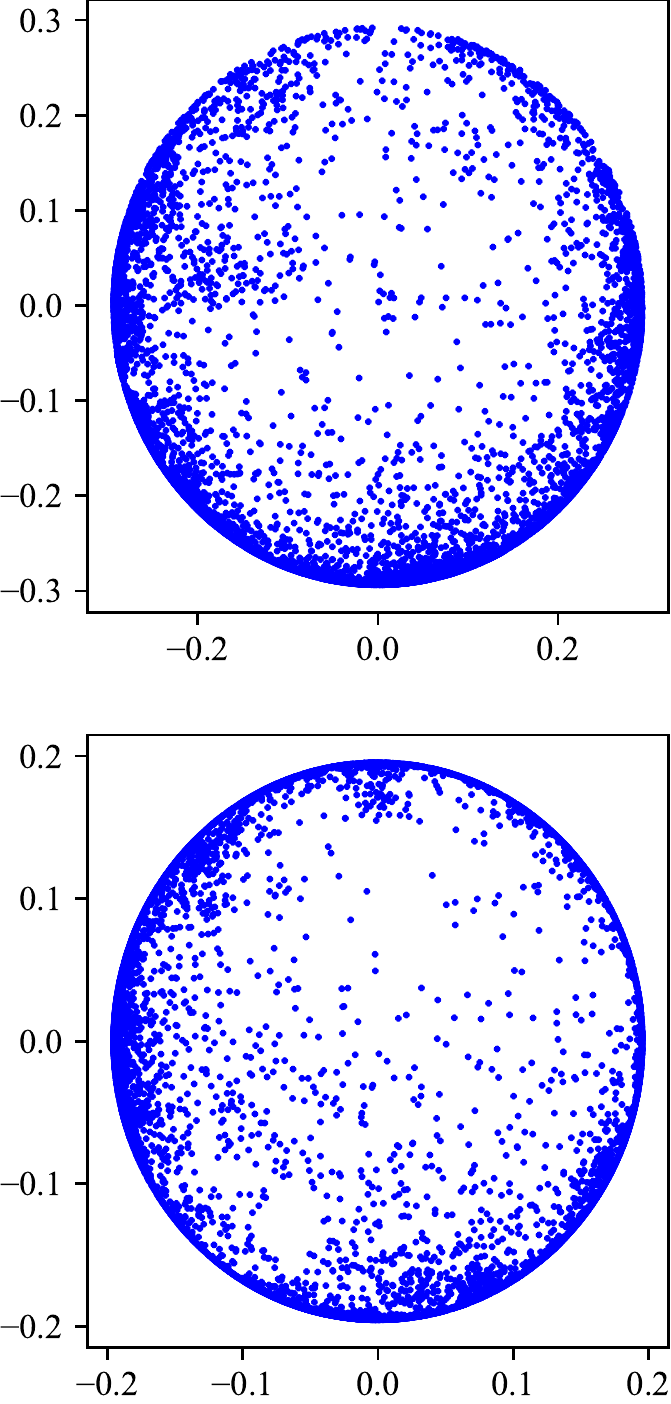}
    \caption{Illustration of two-dimensional hyperbolic embeddings learned by \textsc{AttH} (top) and \textsc{Hercules} (bottom) on ICEWS05-15.}
    \label{fig:2d_hyperbolic_embeddings_atth_icews15}
\end{figure}

We give an illustration of learned embeddings on ICEWS05-15 by \textsc{AttH} and \textsc{Hercules} models in Fig. \ref{fig:2d_hyperbolic_embeddings_atth_icews15}.

\end{document}